\documentclass{article}

\usepackage{PRIMEarxiv}
\usepackage{authblk}
\usepackage[utf8]{inputenc} % allow utf-8 input
\usepackage[T1]{fontenc}    % use 8-bit T1 fonts
\usepackage{hyperref}       % hyperlinks
\usepackage{url}            % simple URL typesetting
\usepackage{booktabs}       % professional-quality tables
\usepackage{amsfonts}       % blackboard math symbols
\usepackage{nicefrac}       % compact symbols for 1/2, etc.
\usepackage{microtype}      % microtypography
\usepackage{lipsum}
\usepackage{fancyhdr}       % header
\usepackage{graphicx}       % graphics
\graphicspath{{media/}}     % organize your images and other figures under media/ folder
\usepackage{multirow}
\usepackage{booktabs}
\usepackage{amsmath,amsfonts}
\usepackage{algorithmic}
\usepackage{algorithm}
\usepackage{array}
\usepackage{url}
\usepackage{caption}
\title{Forgery Guided Learning Strategy with Dual Perception Network for Deepfake Cross-domain Detection}

\author[1]{Lixin Jia}
\author[1,2,*]{Zhiqing Guo} % Added a '*' for the corresponding author example
\author[3]{Gaobo Yang}
\author[1,2]{Liejun Wang}
\author[4]{Keqin Li}

\affil[1]{School of Computer Science and Technology, Xinjiang University, Urumqi, China}
\affil[2]{Xinjiang Multimodal Intelligent Processing and Information Security Engineering Technology Research Center, Urumqi, China}
\affil[3]{College of Computer Science and Electronic Engineering, Hunan University, Changsha, China}
\affil[4]{Department of Computer Science, State University of New York, New Paltz, New York, USA}

\affil[*]{Corresponding author: guozhiqing@xju.edu.cn} 

\makeatletter
\renewcommand{\maketitle}{%
    \begin{center}
    \vspace{2em} 
    {\Large \bfseries \@title \par}
    \vspace{1.5em}
    {\large \@author \par} 
    \vspace{1em}
    \end{center}
}
\makeatother

\begin{document}
\maketitle

\begin{abstract}
The emergence of deepfake technology has introduced a range of societal problems, garnering considerable attention. Current deepfake detection methods perform well on specific datasets, but exhibit poor performance when applied to datasets with unknown forgery techniques. Moreover, as the gap between emerging and traditional forgery techniques continues to widen, cross-domain detection methods that rely on common forgery traces are becoming increasingly ineffective. This situation highlights the urgency of developing deepfake detection technology with strong generalization to cope with fast iterative forgery techniques. 
To address these challenges, we propose a Forgery Guided Learning (FGL) strategy designed to enable detection networks to continuously adapt to unknown forgery techniques. Specifically, the FGL strategy captures the differential information between known and unknown forgery techniques, allowing the model to dynamically adjust its learning process in real time. 
To further improve the ability to perceive forgery traces, we design a Dual Perception Network (DPNet) that captures both differences and relationships among forgery traces. In the frequency stream, the network dynamically perceives and extracts discriminative features across various forgery techniques, establishing essential detection cues. These features are then integrated with spatial features and projected into the embedding space. In addition, graph convolution is employed to perceive relationships across the entire feature space, facilitating a more comprehensive understanding of forgery trace correlations. 
Extensive experiments show that our approach generalizes well across different scenarios and effectively handles unknown forgery challenges, providing robust support for deepfake detection. Our code is available on https://github.com/vpsg-research/FGL.
\end{abstract}

% keywords can be removed

\section{Introduction}\label{sec1}
In the past decade, the extensive application of deep learning has promoted the rapid development of deepfake technology. Various face forgery techniques have emerged, such as NVAE\cite{nvae}, LDM\cite{ldm}. Using these forgery techniques, non-professionals can easily generate forged images and videos. Consequently, numerous forged content was uploaded to mainstream social media platforms, posing significant social risks. With the deepening of the research on Generative Adversarial Networks (GANs)\cite{GAN}, existing deepfake techniques\cite{stylegan3} have continuously been improved. As a result, the face forgery images generated by deepfake are increasingly realistic, which is difficult to distinguish with the naked eye. If misused by malicious actors, deepfake technology could lead to financial fraud, social instability, and even political crises. 

\begin{figure*}[htbp] 
    \centering
    \includegraphics[width=\textwidth]{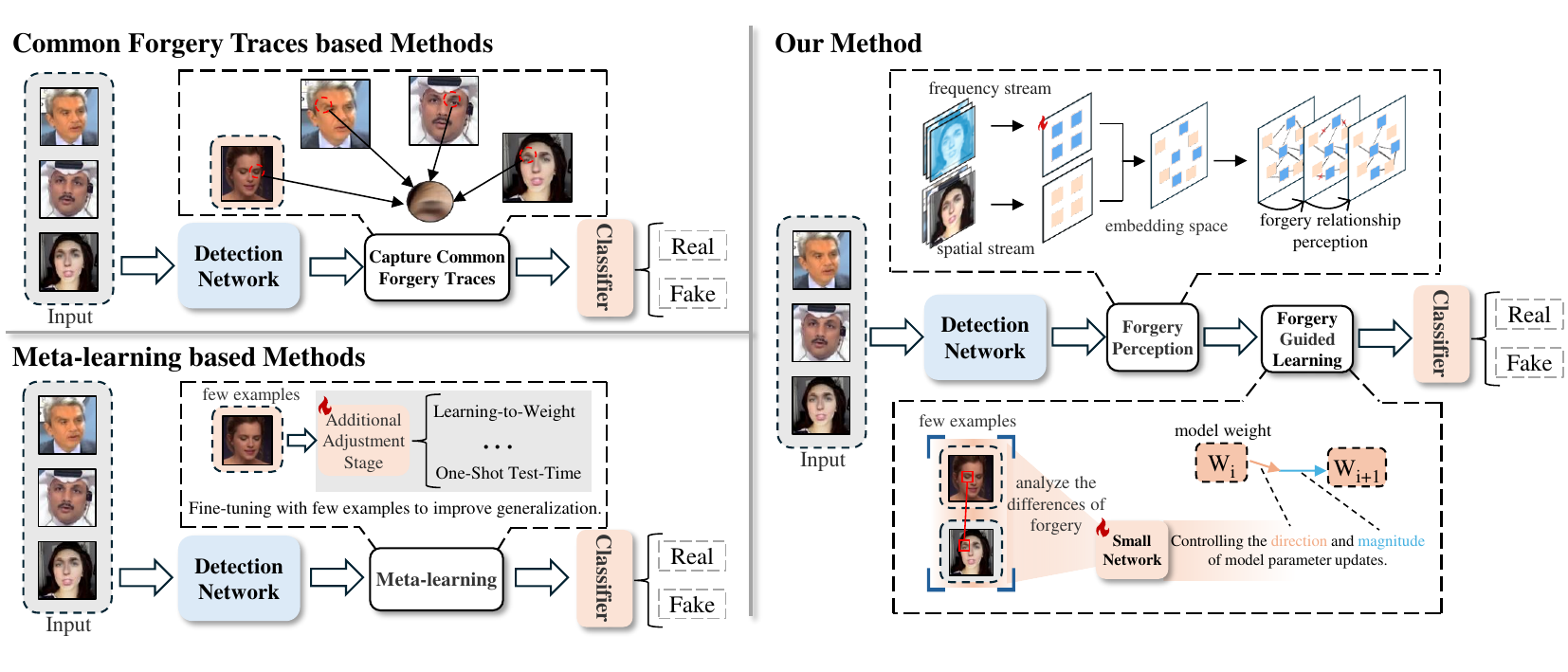} 
    \caption{Comparison with mainstream methods for deepfake cross-domain detection. In the figure, the face images with an orange background represent those generated by unknown forgery techniques. As shown at the bottom, our method analyzes differences between forgery techniques and progressively updates model weights. Additionally, the frequency stream dynamically captures diverse forgery features, which are then integrated with the spatial stream to obtain a more discriminative embedding space. }
    \label{fig:1}
\end{figure*}

At present, various deepfake detection methods \cite{ECCV20, LDFnet} have been proposed to address the problems posed by deepfake technology. These methods achieve high detection performance on public datasets such as FaceForensics++\cite{FF++} and Celeb-DF\cite{CDF}. However, detection models primarily identify forged content by relying on the specific forgery traces present in these datasets. As a result, their effectiveness is limited in real-world scenarios, particularly when confronted with unknown forgery techniques. In the actual environment, new forgery techniques appear constantly and spread rapidly on the internet, which poses a great challenge to effectively detect the content generated by these techniques. 

To overcome this challenge, extensive research\cite{Exposing, F3Net, LSC} has focused on detecting deepfake content by capturing common forgery traces across different techniques. For example, some research has found that deepfake videos often exhibit subtle artifacts or unnatural textures in specific areas\cite{CommonForgery},\cite{Exploit}, such as the edges of the face, eyes, or mouth. These artifacts can serve as important features for distinguishing forged content. 
Motivated by meta-learning\cite{MAML} and few-shot learning\cite{FSL} theories, some deepfake detection algorithms attempt to generalize to unknown forgery domains\cite{LTW},\cite{ost}. 
Although these methods have demonstrated some effectiveness in cross-domain detection and shown potential for detecting unknown forgery techniques, they still exhibit certain limitations. 
(1) As forgery techniques continue to evolve, the gap between known and unknown techniques is widening. The latest forgery model can create high-resolution forged content without obvious artifacts. Consequently, the cross-domain detection methods that rely on identifying common forgery traces will progressively become less effective. 
(2) The distribution of forgery samples across different domains in the feature space is highly inconsistent,  which significantly aggravates the deviation of feature distribution between the source and target domains. Moreover, this deviation hinders the model's ability to capture unknown forgery patterns. 
(3) The meta-learning based methods aim at gradually adapting to the unique characteristics of the new forgery samples, but this usually leads to catastrophic forgetting of prior knowledge. Additionally, the static update strategy struggles to effectively address continuously evolving forgery techniques. 

In this work, we address the limitations mentioned above by introducing the Forgery Guided Learning (FGL) strategy, which enhances the adaptability of detection networks to unknown forgery techniques and improves cross-domain detection performance. As illustrated at the bottom of Fig. \ref{fig:1}, the network capture different forgery information between known and unknown forgery technique through the FGL strategy. By analyzing these differences, the model can dynamically adjust its update direction and magnitude in real time. This strategy is helpful to reduce the forgetting of domain-invariant forgery features and enhance the ability of the model to detect unknown forgery patterns. Due to the discrepancy in feature distributions between known and unknown forged samples, we design a novel Dual Perception Network (DPNet) to perceive both the differences and relationships among forgery features. With its assistance, FGL effectively leverages representative and transferable shared forgery features extracted from known forgery techniques, so that it can learn more general feature representation. 

The specific innovations and contributions of this work are as follows:
\begin{itemize}
\item We propose a novel FGL strategy that adaptively adjusts model parameters by capturing feature differences between known and unknown forgeries. Based on prior knowledge and these feature differences, the strategy provides dynamic feedback on the current loss and gradients, enabling the model to be optimized gradually. 
\item A novel Frequency-domain Perception Mechanism (FPM) is incorporated into the DPNet frequency stream. FPM dynamically captures distinct forgery features in the frequency domain through a dynamic routing mechanism and integrates them with spatial stream features to construct a more discriminative embedding space. 
\item To mitigate feature redundancy and enhance the interaction among forgery features, we integrate graph convolution theory into DPNet. Specifically, the Adaptive Forgery Relationship Perception (AFRP) module learns feature representations in the embedding space, capturing the relationships between forged and non-forged features while dynamically adjusting them. 
\item We conducted extensive experiments across five datasets, simulating various real-world scenarios, to comprehensively validate the effectiveness of our method. The results demonstrate that the proposed approach can significantly improve the detection of unknown forgery techniques with a few examples, highlighting its significant application value. 
\end{itemize}

The remainder of this paper is organized as follows. Section \ref{sec2} reviews related work on deepfake detection techniques. Section \ref{sec3} presents our proposed approach. Section \ref{sec4} reports the evaluation results across different experimental scenarios. Section \ref{sec5} concludes the paper and discusses its limitations and future directions.

\section{RELATED WORKS}\label{sec2}
\subsection{Spatial-based and Frequency-based Deepfake Detection}
In recent years, many face forgery detection methods have been developed to mitigate the risks associated with the misuse of deepfake technology. Early detection methods primarily focus on identifying localized artifacts and inconsistencies in forged images. For example, the Face X-ray method\cite{facex-ray} highlights the presence of blending discrepancies between forged facial regions and background images, which serve as distinctive indicators for detecting manipulations. Li et al. \cite{li2018exposing} demonstrated that deepfake videos often exhibit unique visual artifacts, which can be effectively captured using Convolutional Neural Networks (CNN) to distinguish the authenticity of face images. These findings have significantly advanced the development of deepfake detection techniques. 

\begin{figure}[htbp]
    \centering
    \includegraphics[width=0.7\columnwidth]{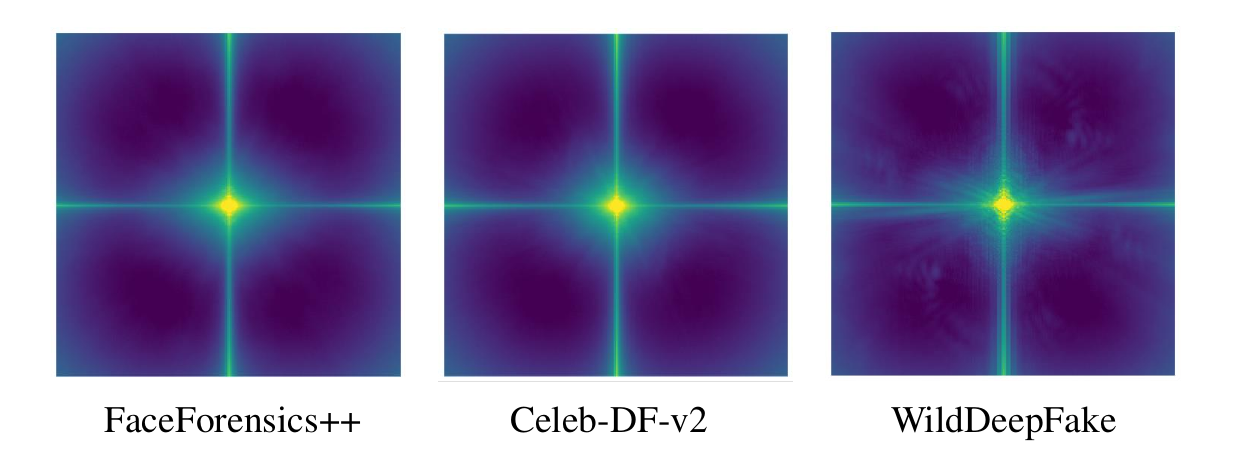} % 设置图片宽度为 \textwidth（页面宽度）
    \caption{Frequency analysis of various deepfake datasets. Each FFT spectrum plot represents the average result of 2,000 images randomly selected from the corresponding dataset. }
    \label{fig:6}
\end{figure}

In addition to analyzing forgery traces in the spatial domain, researchers have identified significant distinctions between fake and real images in the frequency domain. F3Net \cite{F3Net} integrates information in the frequency domain to CNN, allowing the effective detection of subtle manipulation traces in forged facial images. 
However, recent studies \cite{DiffFrequency} reveal that different forgery techniques produce distinct frequency domain artifacts. As illustrated in Fig. \ref{fig:6}, we perform a frequency spectrum analysis of various deepfake datasets using the Fast Fourier Transform (FFT), revealing subtle differences across datasets.  
Inspired by these observation, we propose the Frequency-domain Perception Mechanism (FPM). Unlike existing methods that rely on static frequency-domain forgery cues, our FPM progressively captures sample-specific forgery cues in the frequency domain. By integrating these dynamic frequency with spatial information, we construct a more discriminative embedding space, thereby enhancing both the accuracy and robustness of forgery detection.

\subsection{Graph-based Deepfake Detection}
An emerging trend is the application of Graph Neural Networks (GNN) in the field of computer vision. Vision GNN\cite{Visiongnn} proposed a new method to represent an image as a graph structure. By dividing the image into blocks as nodes and combining graph convolution operation to construct adjacency relations, the issue of over-smoothing in traditional GNN is addressed. At the same time, GNN provides a new approach to improve the generalization ability of deepfake detection. 
Yang et al.\cite{yangmasked} regarded deepfake detection as a graph classification problem, where each face region corresponds to a vertex, offering a new perspective. At the same time, they found that a large amount of redundant relationship information in the graph would hinder the expressive ability of the graph, so they proposed a masking relationship learning method to reduce redundant information.

Similarly, the redundancy of forgery feature information can hinder the discriminative accuracy of detection networks. To address this issue, we introduce the Adaptive Forgery Relationship Perception (AFRP) into the detection network, forming a Dual Perception Network together with the FPM. The AFRP treats features in the embedding space as nodes and utilizes an adjacency matrix to perceive and dynamically adjust their relationships. 
Different from the masked relational learning method in reference \cite{yangmasked}, which mainly reduces redundancy, AFRP focus on enhancing feature representation. This allows the model to better attend to semantically relevant and informative features. 

\subsection{Generalization-based Deepfake Detection}
As more and more deepfake datasets and forgery techniques emerge, achieving effective cross-domain detection has become a critical research challenge. Therefore, numerous studies\cite{luo2023beyond},\cite{inconsistency} have proposed different methods to improve the generalization ability of the detection model. For example, Zhao et al.\cite{MAT} proposed a multi-attention deepfake detection network that captures subtle differences between real and fake images. DFGaze\cite{DFGaze} analyzes gaze features in facial video frames to more effectively identify cues between real and fake faces. Dong et al.\cite{CADDM} proposed an ID-unaware deepfake detection model, which enhanced the generalization ability by reducing the learning of image identity information. Although these methods perform well in handling common forgery techniques, their effectiveness is limited when confronted with unknown forgeries that lack obvious shared features. 

Other studies have proposed novel approaches to address emerging forgery techniques. By combining meta-learning technology, Sun et al.\cite{LTW} designed a Learning-to-Weight framework. They argue that different faces contribute differently to the model across multiple domains, which can cause model deviation in specific domains. During the meta-test stage, the loss function is employed to guide the model update, selecting the model that is most suitable for the target domain. Ni et al.\cite{CORE} proposed a Consistent Representation Learning method, which captures diverse representations through different augmentations and then applies regularization to enhance consistency in forgery detection. 
In another study based on meta-learning\cite{ost}, it improves model adaptability by updating the model before applying it to the test samples. This is achieved by constructing auxiliary tasks specific to the test samples. 

However, the static update strategy adopted by these detection methods shows an obvious forgetting phenomenon in the incremental learning process, which eventually leads to the decline of the generalization ability of the model. 
In this paper, we propose a new solution called Forgery Guided Learning (FGL) strategy. The FGL guides model updating incrementally by learning the feature differences between forgery samples from the source and target domains. Additionally, we adaptively adjust the direction and magnitude of each update step based on the parameters and gradient of the current model, thus improving the adaptability and generalization ability of the model. 

\begin{figure*}[htbp] % 使用 figure* 横跨两栏
    \centering
    \includegraphics[width=\textwidth]{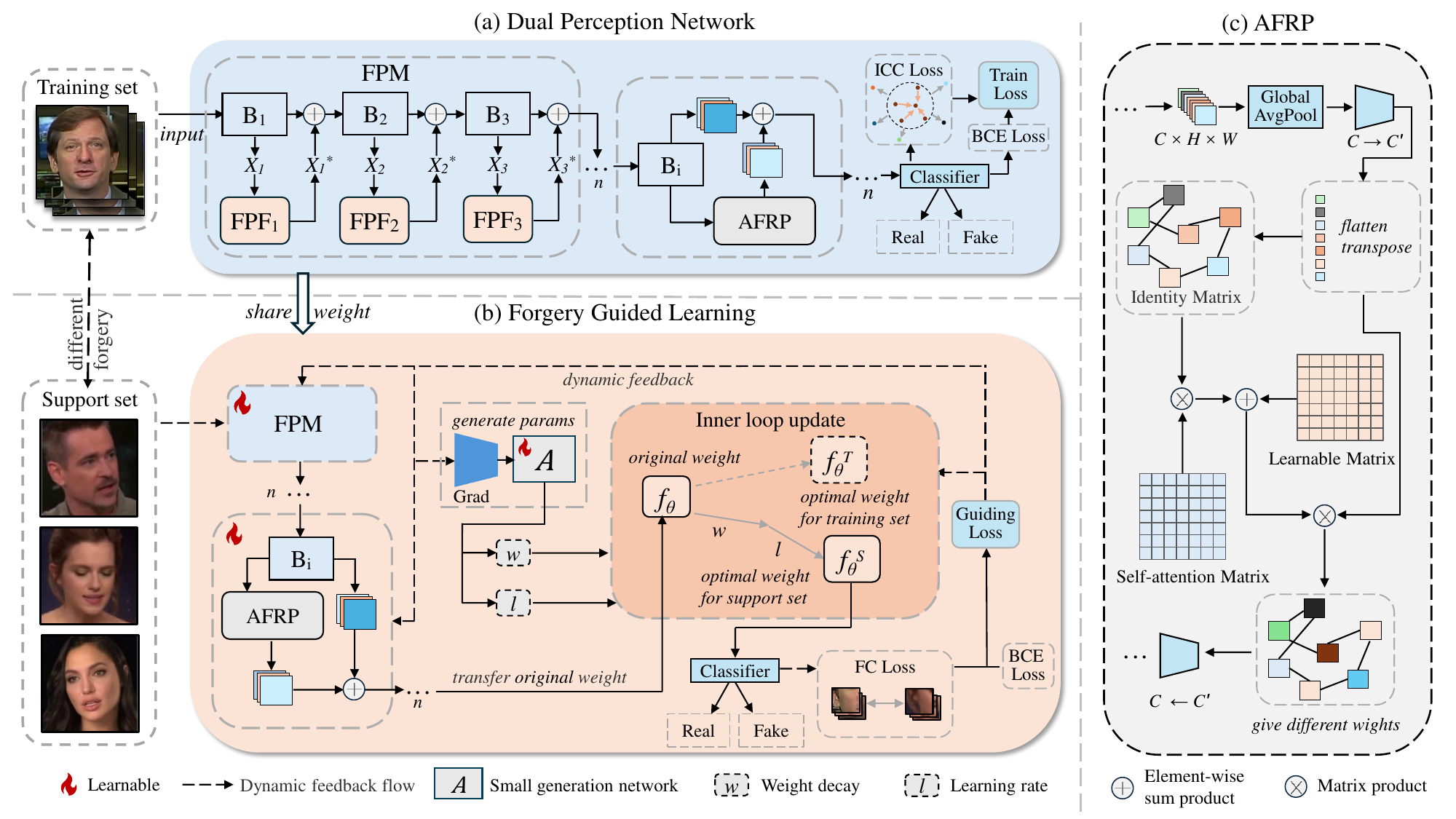} % 设置图片宽度为 \textwidth（页面宽度）
    \caption{The overall architecture of our proposed method. Figure (a) illustrates the Dual Perception Network (DPNet). The training set is fed into the model, and dynamic frequency-domain information is introduced through the Frequency-domain Perception Mechanism (FPM). Then the key information are distinguished by multiple Adaptive Forgery Relationship Perception (AFRP). Finally, ICC Loss assisted BCE Loss to guide model learning. Figure (b) represents the Forgery Guided Learning (FGL) strategy. In this phase, this strategy is proposed to guide the detection model in improving its performance on unknown forgery techniques. FGL adjusts the model parameters by learning the differences between the support set and the original dataset. The loss in this phase is composed of both the FC Loss and BCE Loss. Figure (c) provides a detailed illustration of AFRP.}
    \label{fig:2}
\end{figure*}

\section{PROPOSED APPROACH}\label{sec3}

In this section, we introduce the proposed method. The overall framework is shown in Fig. \ref{fig:2}. 
We treat the original dataset as the training set ($\mathcal{T}$) and the few examples as the support set ($\mathcal{S}$). Our framework is structured into two stages. In the first stage, the Dual Perception Network (DPNet) is trained on the $\mathcal{T}$ to learn fundamental forgery patterns. The Frequency-domain Perception Mechanism (FPM) incrementally extracts forgery cues across multiple levels in the frequency domain, capturing forgery features at different scales. Subsequently, the Adaptive Forgery Relationship Perception (AFRP) module analyzes the relationships among these forgery features and dynamically adapts based on their variations. 
In the second stage, we propose a Forgery Guided Learning (FGL) strategy, which uses a few support sets of unknown forgery techniques to guide the model's learning process. 
The following sections provide a detailed introduction to our approach.

\subsection{Forgery Guided Learning Strategy}
To enhance the adaptability of the detection model to unknown forgery techniques, we introduce the Forgery Guided Learning (FGL) strategy, which enables the model to effectively generalize across diverse forgery patterns. The entire FGL strategy is illustrated in Fig. \ref{fig:2} (b). We incorporate the fundamental forgery patterns learned from the Dual Perception Network (DPNet) as prior knowledge, denoted as $f_{\theta}$, which represents the weight parameters of the entire model. Subsequently, $f_{\theta}$ is progressively updated using the FGL strategy and is evaluated using the corresponding loss function $L_{\mathcal{S}}^{\mathcal{T}}$ to assess its generalization ability to unknown forgery techniques. 

Traditional detection methods typically aim to find an optimal weight that achieves good generalization across different forgery techniques. But the weight obtained in this way is often the optimal weight $f_{\theta}^{\mathcal{T}}$ for the training set. Existing research has shown that designing a better rapid adaptation method can improve the adaptability to tasks when learning with few samples\cite{FSL}. Inspired by this view, our approach focuses on adapting to unknown forgery domains through the FGL strategy to obtain the optimal weights $f_{\theta}^{\mathcal{S}}$ for unknown forgery techniques. In order to prevent the model from overfitting the few samples $\mathcal{S}$, we introduce a $l_2$ regularization term $\frac{\lambda}{2}||\theta||_2$. The basic parameter update in our FGL strategy is presented in Equation \eqref{1}.
\begin{align}
\label{1}
\theta _{i+1} &= \theta _{i} - l(\nabla_\theta  L_{\mathcal{S}}^{\mathcal{T}}(f_{\theta _{i}}) + \lambda \theta _{i}) \nonumber \\
              &= w \theta _{i} - l \nabla_\theta  L_{\mathcal{S}}^{\mathcal{T}}(f_{\theta_{i}}),
\end{align}
where, $i$ represents the step index, $f_{ \theta }$ represents the model weight parameters from DPNet, and $\theta$ represents the updated weight parameters. The two control variables, $l$ and $w$, control the direction and magnitude of updates of the model parameters. Essentially, the learning rate and weight update amount are determined by the loss and gradient computed from the current sample, denoted as $l$. On the other hand, $w$ controls the degree of regularization, specifically determining the extent to which the existing knowledge should undergo weight decay.

These two control variables alone are not enough to face all possible forgery detection situations, so we need to have more flexibility when updating the model parameters. Therefore, two control variables can be replaced with adjustable variables $l_{i}$ and $w_{i}$, which have the same dimensions as $\nabla_\theta  L_{\mathcal{S}}^{\mathcal{T}}(f_{\theta_{i}})$ and $\theta _{i}$. Finally, our FGL strategy equation becomes
\begin{equation}
\label{2}
\theta_{i+1}=w_{i}\odot\theta_{i}-l_{i}\odot\nabla_{\theta}L_{\mathcal{S}}^{\mathcal{T}}(f_{{\theta}_{i}}),
\end{equation}
where $\odot$ denotes element-wise multiplication. The purpose of the FGL strategy is to gradually guide the model's adaptation to unknown forgery techniques. To guide the model learning process more precisely, we generate control variables at step $i$ based on the specific learning state $t_{i}$. In the proposed framework, the control variables $l_{i}$ and $w_{i}$ are generated by a small generation network $\mathcal{A}$, as follows:
\begin{equation}
\label{3}
(l_{i},w_{i})=\mathcal{A}(t_{i}).
\end{equation}

As shown in Fig. \ref{fig:2} (b), the small generation network $\mathcal{A}$ generates specific control variables before each inner loop update. We construct $\mathcal{A}$ using a two-layer MLP, incorporating a ReLU activation function between the layers. The $\mathcal{A}$ takes a vector $t_{i}$ as input, which is twice the number of layers in the base model DPNet. The outputs $l_{i}$ and $w_{i}$ are generated layer by layer and then expanded to match the dimensions of their respective parameters $\theta_{i}$. These control variables are then used as the learning rate and weight decay factor to adjust the direction and magnitude of parameter updates during each step of the inner loop.

During the process of guiding the detection model's learning, the model's updates rely on the current learning state of each step. We define the learning state $t_{i}$ as consisting of the current model weights $\theta_{i}$ and the corresponding gradient $\delta_{i}=\nabla_{{\theta}}L_{\mathcal{S}}^{\mathcal{T}}(f_{\boldsymbol{\theta}_{i}})$. In order to calculate the hierarchical average of weights and gradients, we calculate the average of all layers of base model as follows:
\begin{equation}
\label{4}
\boldsymbol{\bar{\theta}_{i}}=\frac1n\sum_{l=1}^{n}{\theta}_{i}^{l},\quad\boldsymbol{\bar{\delta}_{i}}=\frac1n\sum_{l=1}^{n}{\delta}_{i}^{l},
\end{equation}
where $l$ represents the number of layers in the model, $\boldsymbol{\bar{\theta}_{i}}$ and $\boldsymbol{\bar{\delta}_{i}}$ represent the averaged weights and gradients, respectively. This leads to the following learning state:
\begin{equation}
\label{5}
t_{i}=[\boldsymbol{\bar{\theta}_{i}},\boldsymbol{\bar{\delta}_{i}}].
\end{equation}

The learning state $t_{i}$ not only captures the weight and gradient distribution of the current model, but also reflects the direction and speed of the guided learning process through its dynamic changes. Generally speaking, in the initial stage of FGL, the model parameters are updated more and the gradient is larger, and finally it will tend to be stable.

\begin{figure}[htbp] 
    \centering
    \includegraphics[width=0.6\columnwidth]{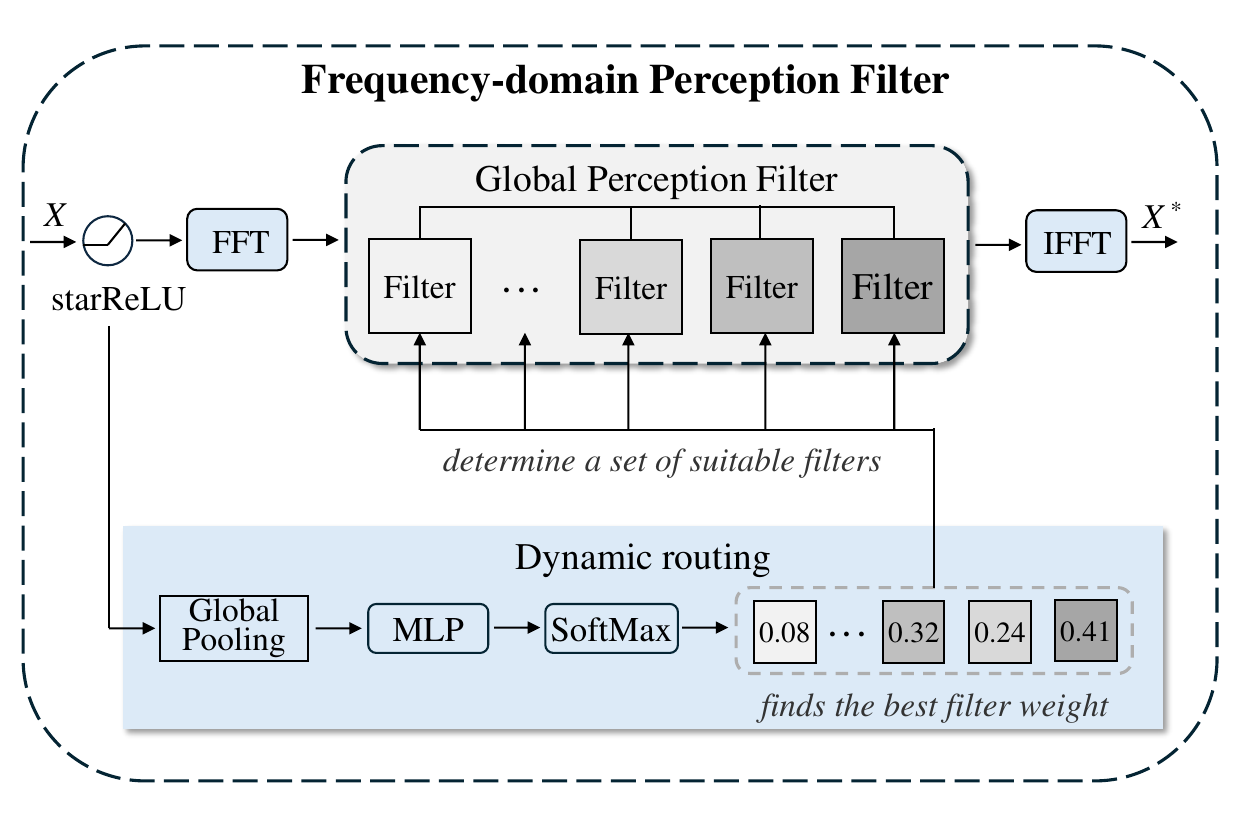} 
    \caption{Detail structure of FPF. We use a dynamic routing mechanism to find the most suitable filter weights for the current input. These weights are then assigned to different filters to dynamically determine a globally perception filter. Additionally, we use starReLU as our activation function.}
    \label{fig:3}
\end{figure}

\subsection{Dual Perception Network}

\subsubsection{Frequency-domain Perception Mechanism}

Since different forgery techniques will expose different characteristics in frequency domain, we design a Frequency-domain Perception Mechanism (FPM) as shown in the left of Fig. \ref{fig:2} (a). Inspired by dynamic filter\cite{dynamicfilter}, we use a set of Frequency-domain Perception Filter (FPF) to adjust the extracted multi-level information dynamically, introducing frequency-domain features from different levels. As shown in Fig. \ref{fig:3}, FPF dynamically determines a global perception filter using multiple filters. The idea of FPF can be simply described as follows:
\begin{equation}
\label{6}
X^*=\mathcal{F}^{-1}\left(\sum_{k=1}^Kw_k(X)\cdot\left(W_f^{(k)}\odot\mathcal{F}(\mathcal{R}(X))\right)\right),
\end{equation}
where $X$ represents the input feature tensor, and $X^*$ is the final output feature carrying dynamic frequency-domain information, and $\odot$ denotes element-wise multiplication. We use $\mathcal{F}$ and $\mathcal{F}^{-1}$ to represent the Fast Fourier Transform and its inverse transform. $W_f^{(k)}$ represents the $k$-th frequency-domain filter. The filter's weight $w_k(X)$ is generated by a dynamic routing mechanism\cite{Dynamicrouting}. $\mathcal{R}(X)$ is the activation function for the input. Here, we use the starReLU activation function proposed by Yu\cite{starReLU}, which is defined as follows: 
\begin{equation}
\label{7}
\mathcal{R}(X)=s\cdot\mathrm{ReLU}(X)^2+b.
\end{equation}
By using a learnable scale $s$ and a bias $b$, the activation function can be dynamically adjusted during training, enhancing the flexibility of the method.

Dynamic routing mechanism adaptively finds the best filter weight in frequency-domain through input features. Based on the characteristics of the input data, we allocate weights among different filters to dynamically select a globally perception filter that best suits the current input. Specifically, the global features of the input are first extracted by global pooling $\mathrm{G}(X)$ to describe the overall information of the whole input feature. Then, the global features are mapped to the dynamic weight space by a lightweight Multilayer Perceptron (MLP). Finally, the process of generating the weight $w_k(X)$ is as follows:
\begin{equation}
\label{8}
w_k(X)=\frac{exp\left(\phi_k\cdot\mathrm{MLP}(\mathrm{G}(X))\right)}{\sum_{j=1}^Kexp\left(\phi_j\cdot\mathrm{MLP}(\mathrm{G}(X))\right)},
\end{equation}
$\phi_k$ is a learnable scalar, which is optimized by training and used to measure the importance of each filter. Where $exp$ and the normalization term ensure that the generated weights $w_k(X)$ satisfy a probability distribution $\sum_{k=1}^Kw_k(X)=1$.

By introducing $\phi_k$ during weight generation, the process not only depends on the input features, but also adaptively adjusts the initial importance of each filter. The dynamically generated weights $w_k(X)$ are used to weight the filter $W_f^{(k)}$, and then element-wise multiplied with the frequency-domain feature $X_\mathcal{F}$, 
\begin{equation}
\label{9}
X_w=\sum_{k=1}^Kw_k(X)\cdot\left(W_f^{(k)}\odot X_\mathcal{F}\right).
\end{equation}
The resulting $X_w$ incorporates dynamic frequency-domain information, which aids the detection model in identifying distinctive forgery traces in the frequency domain. These features are then integrated with forgery features in the spatial domain and projected in the embedding space.

\subsubsection{Adaptive Forgery Relationship Perception}
After incorporating dynamic frequency domain information, the representation of forged features in the embedding space is further enriched. However, certain redundant or irrelevant information may negatively impact the model's performance. Additionally, adaptive graph convolution \cite{agca} has been shown to focus on semantically relevant and informative features. To address this issue, we introduce the Adaptive Forgery Relationship Perception (AFRP) module and integrate it into the Dual Perception Network (DPNet) to facilitate the extraction of salient information, as illustrated in Fig. \ref{fig:2} (c). The core concept of AFRP is to model features as graph vertices in embedded space, and use graph convolution theory to establish their dependencies, so as to realize a more structured and differentiated representation of forged features. 

First, we will process the input feature map. Specifically, the input feature map $x\in\mathbb{R}^{B\times C\times H\times W}$ is passed through the global average pooling $\mathrm{G}(x)$, then flattened and transposed into a feature vertex matrix. The processing of the feature map is expressed as follows:
\begin{equation}
\label{10}
\begin{cases}V^{\prime}=\mathrm{Conv}_r\left(\mathrm{G}(X)\right),\quad V^{\prime}\in\mathbb{R}^{B\times C^{\prime}\times1\times1} &\\ V=\mathrm{Transpose}\left(\mathrm{Flatten}(V^{\prime}\right)),\quad V\in\mathbb{R}^{B\times1\times C^{\prime}} & \end{cases}
\end{equation}
where $C^{\prime}=\frac{C}{r}$, we use $\mathrm{Conv}_r $ to reduce the dimensionality of the feature map. To reduce computational complexity, we introduce a bottleneck structure that reduces the feature dimension from $C$ to $\frac{C}{r}$, and then restores it back. This operation also improves the efficiency of the interaction between features. 

To describe the dependencies between the feature vertices, the AFRP designs an adjacency matrix consisting of three components. First, the identity matrix $M_I=I$ represents the self-connection relationships between the feature vertices. Next, the self-attention matrix $M_{sa}$, is used to emphasize the weight of each feature vertex. Finally, the learnable global adjacency matrix $M_l$ captures the global relationships between any two feature vertices. The final adjacency matrix $M$ is obtained as follows:
\begin{equation}
\label{11}
\begin{cases}\begin{aligned}&M=M_I\odot M_{sa}+M_l,\\&M_I=I,\quad M_I\in\mathbb{R}^{C^{\prime}\times C^{\prime}}\\&M_{sa}=T\left(S\left(\mathrm{Conv}(V)\right)\right),\quad M_{sa}\in\mathbb{R}^{C^{\prime}\times C^{\prime}}\\&M_l=\mathrm{Parameter}(1\times10^{-6}),\quad M_l\in\mathbb{R}^{C^{\prime}\times C^{\prime}}\end{aligned} \end{cases}
\end{equation}
where $S$ stands for Softmax function and $T$ stands for arranging the results as a diagonal matrix. Parameter refers to the learnable parameters that are updated during model gradient descent. We initialize the parameter to $1\times10^{-6}$ to ensure the stability of learning.

Finally, the generated adjacency matrix $M$ is used for graph convolution operation, so as to adjust the importance of features. And restore the feature dimension to $X^{\prime}\in\mathbb{R}^{B\times C\times H\times W}$ through $\mathrm{Conv_{ir}}$. The whole process can be expressed as:
\begin{equation}
\label{12}
\begin{cases}X^{\prime}=X\cdot\sigma\left(\mathrm{Conv_{ir}}\left(\mathrm{ReLU}\left(\mathrm{Conv}\left(V\cdot A\right)\right)\right)\right),&\\X^{\prime}\in\mathbb{R}^{B\times C\times H\times W},\end{cases}
\end{equation}
where $\sigma$ indicates that the Sigmoid activation function is used to normalize the output weights. By learning the weight relationships of these features through AFRP, the model enhances useful features on a global scale while suppressing irrelevant ones. This global feature processing approach not only improves the model's adaptability to unknown forgery techniques but also enhances its performance on unknown datasets.

\subsection{Loss Function}
Since DPNet and FGL strategy are designed to achieve different goals, independent loss functions are adopted for each part to ensure optimal learning and performance. For the loss function of DPNet, we use the Intra-class Compact (ICC) loss proposed by Sun et al. \cite{LTW} to assist the Binary Cross-entropy (BCE) loss for training. The basic idea of ICC loss is to gather positive samples and keep all negative samples away from the real center $C_{real}$. By adopting this loss, we can promote the model to explore more discriminant features and improve the generalization of the model, which is in line with our research ideas. ICC Loss is defined as: 
\begin{equation}
\label{13}
\begin{cases}L_{icc}=L_{positive}-L_{negative},&\\
L_{positive}=\frac{1}{|O^{real}|}\sum_{j=1}^{|O^{real}|}(o_{j}^{real}-C_{real})^{2},&\\
L_{negative}=\frac{1}{|O^{fake}|}\sum_{j=1}^{|O^{fake}|}(o_{j}^{fake}-C_{real})^{2},
\end{cases}
\end{equation}
where $O^{real}$ and $O^{fake}$ represent the sets of positive and negative samples respectively. Then use $\lambda$ to combine the weighted ICC loss with the BCE loss to get the final loss:
\begin{equation}
\label{14}
L_{\mathcal{T}}=L_{bce}+\lambda L_{icc}.
\end{equation}

For the loss function of FGL, we draw inspiration from the InfoNCE loss \cite{Loss} and integrate the Feature Contrastive (FC) loss into our objective function to enhance feature discrimination and learning efficiency. Firstly, the feature $X\in\mathbb{R}^{B\times C\times H\times W}$ extracted from the model is flattened into $F\in\mathbb{R}^{B\times D}$ and the similarity matrix between them is calculated:
\begin{equation}
\label{15}
{S}_{i,j}=\frac{{X}_i\cdot{X}_j}{\|{X}_i\|\|{X}_j\|\cdot\tau}, \quad{S}_{i,j}\in\mathbb{R}^{B\times B}
\end{equation}
where $\tau$ is a temperature parameter that controls the smoothness of the similarity distribution. Finally the formula for the FC loss is given as: 
\begin{equation}
\label{16}
{L}_{fc}=-\frac{1}{B}\sum_{i=1}^B\log\frac{exp({S}_{i,j})}{\sum_{j=1}^Bexp({S}_{i,j})}.
\end{equation}

In FGL, parameter updates rely on task gradients computed by the model. The FC loss provides additional gradient information from feature similarity, which differs from classification loss. By introducing contrastive loss, the diversity of gradients is enhanced. We also compute losses from both the source domain and the target domain. The overall loss at this stage is:
\begin{equation}
\label{17}
L_{\mathcal{S}}^{\mathcal{T}}=L_{bce}^{\mathcal{T}}+\mu L_{bce}^{\mathcal{S}}+\nu L_{fc}.
\end{equation}

\section{EXPERIMENTS}
\label{sec4}
To evaluate the effectiveness of our proposed method, we simulate real-world scenarios and conduct experiments on five public deepfake detection datasets, focusing on the following two practical scenarios. (1) In the first scenario, we utilize the widely adopted FaceForensics++ (FF++) dataset \cite{FF++} as the training set. By evaluating the performance on different datasets, we assess the generalization capability of our method when applied to unknown datasets. 
(2) In the second scenario, we conduct experiments on the FF++ dataset, where three out of four forgery techniques are used for training, while the remaining unseen forgery method is used for testing. This setup allows us to evaluate the effectiveness of our proposed method in handling previously unknown forgery techniques.

\subsection{Datasets}

{\bf{FaceForensics++ (FF++)}} \cite{FF++} is widely regarded as the most commonly used dataset in this field. It offers several benefits, including a large volume of data, a diverse range of forgery techniques, and video samples grouped by varying compression qualities. The dataset includes 1,000 original real videos, with 1,000 corresponding forgery videos generated using four different forgery techniques: Deepfakes (DF) \cite{DF}, Face2Face (F2F) \cite{F2F}, FaceSwap (FS) \cite{FS}, and NeuralTextures (NT) \cite{NT}.

{\bf{Celeb-DF-v2 (CDF)}} \cite{CDF} dataset is a high-quality deepfake dataset tailored for celebrity faces, providing exceptional visual realism. It contains 590 authentic videos and 5,639 corresponding forgery videos.

{\bf{DeeperForensics-1.0 (DFR)}} \cite{DFR} dataset comprises 50,000 real videos and 10,000 forged videos. It not only offers a large volume of data but also incorporates diverse postures, lighting conditions, and expressions, making it more representative of real-world scenarios.

{\bf{WildDeepFake (WDF)}} \cite{WDF} dataset consists of 3,805 real face videos and 3,509 forged face videos. Both real and forged videos are sourced from the internet, providing a diverse range of content that enhances the challenge of detection. 

{\bf{DeepFake Detection Challenge (DFDC)}} \cite{DFDC} constructed the large-scale DFDC dataset, which contains 23,654 real videos captured by 3,426 hired actors in various environments. It also includes 104,500 forged videos created using more advanced deepfake techniques.

\renewcommand{\arraystretch}{1.3}
\begin{table}[t]
\small
\centering
\caption{INTRA-DATASET EVALUATION RESULTS ON FF++}
\label{tab:table1}

\begin{tabular}{cccccc}
\toprule
\multirow{2}{*}{Method} & \multirow{2}{*}{Venue} & \multicolumn{2}{c}{FF++(c23)}   & \multirow{2}{*}{Params.} & \multirow{2}{*}{FLOPS} \\ \cline{3-4}
                        &                        & ACC(\%)            & AUC(\%)            &                          &                        \\ \hline
Xception\cite{FF++}             & ICCV 2019              & 75.60          & 88.63          & 20.81M                   & 6.00G                  \\
F3-Net\cite{F3Net}                  & ECCV 2020              & 93.15          & 96.74          & 21.17M                   & 8.49G                \\
LTW\cite{LTW}                     & AAAI 2021              & 94.31          & 97.80          &20.37M                          &4.20G                        \\
MAT\cite{MAT}                     & CVPR 2021              & 93.72          & 97.26          & 47.88M                   & 5.18G                  \\
RECCE\cite{RECCE}                   & CVPR 2022              & \underline{95.26}    & \textbf{98.22}    & 23.81M                   & 6.17G                  \\
CORE\cite{CORE}                    & CVPR 2022              & 91.75          & 96.99          & 20.81M                   & 4.60G                  \\
CADDM\cite{CADDM}                    & CVPR 2023              & 93.96          & 97.45          &-                    &4.83G                   \\
IFFD\cite{IFFD}                    & TIP 2023               & 94.06          & 97.39          & 20.81M                   & 4.59G                  \\
UMFC\cite{UMFC}                    & AAAI 2024              & 91.61          & 95.48          & 54.01M                   & 9.16G                 \\
SFIConv\cite{SFIConv}                 & TIFS 2024              & 87.53          & 93.57          & 13.95M                   & 3.09G                  \\
DFGaze\cite{DFGaze}                  & TIFS 2024              & 89.44          & 90.90          & 8.54M                   & 3.32G                 \\ \hline
FGL                    & Ours                   & \textbf{95.38} & \underline{98.11}
&20.19M                          &4.80G                        \\ \bottomrule 
\end{tabular}
% }
\end{table}

\subsection{Experimental Settings}
\subsubsection{Implementation Details} For all datasets, we follow the official documentation to divide the datasets. We sample 20 frames from each video to construct our experimental dataset. We use MTCNN \cite{MTCNN} to crop facial regions and resize them to $224\times224$. All experiments are carried out under the framework of PyTorch using a single NVIDIA RTX4090D. The model is initialized with pre-trained parameters from ImageNet-1K \cite{imagenet}. We use the Adam optimizer with parameters $\alpha$ and $\beta$ set to 0.999 and 0.99, respectively. The initial learning rate $L_{r}$ is set to $2\times10^{-4}$ and decays by a factor of 0.5 every 5 epochs. We train each detection model for 20 epochs with a batch size of 32. In the training stage, we also apply data augmentation techniques to enhance data diversity.

\subsubsection{Evaluation Metrics} We follow the current mainstream evaluation strategies, using Accuracy (ACC) and the Area Under the Receiver Operating Characteristic Curve (AUC) as the evaluation metrics for our experiments. Additionally, we evaluate the model's complexity by measuring the number of parameters (Param.) and the floating-point operations (FLOPs).

\subsection{Intra-dataset Evaluation}
We compare our proposed method with the latest and representative methods, including XceptionNet\cite{FF++}, F3-Net\cite{F3Net}, LTW\cite{LTW}, MAT\cite{MAT}, RECCE\cite{RECCE}, CORE\cite{CORE}, CADDM\cite{CADDM}, IFFD\cite{IFFD}, UMFC\cite{UMFC}, SFIConv\cite{SFIConv}, DFGaze\cite{DFGaze}. To ensure a fair comparison, we reproduce the results using the code provided by the authors. Additionally, the datasets and experimental settings used in the experiments are consistent with those of our method. All experimental results are evaluated using image-level metrics, which are more challenging for detection methods. 

First, we evaluate our method on the FF++ dataset, as shown in Table~\ref{tab:table1}. The best results are highlighted in bold, and the second-best results are underlined. Among many excellent methods, our method and RECCE\cite{RECCE} simultaneously achieve the best performance in the FF++ dataset. Compared to the LTW method \cite{LTW}, which is based on meta-learning, our approach achieves superior performance within the dataset, as the Dual Perception Network (DPNet) enables more comprehensive information perception. In general, a detector's performance within a dataset is related to its ability to extract useful forgery feature information from the dataset. Most deepfake detection methods have achieved good performance within the dataset. The RECCE\cite{RECCE} method further achieves an AUC of 98.22\% through its Reconstruction-Classification learning. The advantage of our method is that it can adaptively judge the importance of forgery feature information and finally achieve the highest ACC performance. F3-Net\cite{F3Net} also uses frequency-domain forgery information. In contrast, our method leverages the Frequency-domain Perception Mechanism (FPM) to capture distinct forgery patterns, enabling more effective utilization of frequency-domain information and significantly enhancing model performance. The other detection methods, such as DFGaze\cite{DFGaze} and SFIConv\cite{SFIConv}, put more emphasis on the generalization performance of the detection model, so the performance in the data set is relatively general. For DFGaze\cite{DFGaze}, we maintained identical experimental settings to other models and trained it for 20 epochs, instead of the 100 epochs used in the original paper. 
\renewcommand{\arraystretch}{1.3}
\begin{table*}[thbp]
\small
\setlength{\tabcolsep}{1mm}
\centering
\caption{THE EVALUATION RESULTS ON UNKNOWN FORGERY DATASETS}
\label{tab:table2}

\begin{tabular}{ccccccccccc}
\toprule
\multirow{2}{*}{Method} & \multirow{2}{*}{Venue} & \multicolumn{2}{c}{CDF}         & \multicolumn{2}{c}{DFR}         & \multicolumn{2}{c}{WDF}         & \multicolumn{2}{c}{DFDC}        & \multirow{2}{*}{\begin{tabular}[c]{@{}c@{}}Average\\ AUC(\%)\end{tabular}} \\ \cline{3-10} 
                        &                        & ACC(\%)        & AUC(\%)        & ACC(\%)        & AUC(\%)        & ACC(\%)        & AUC(\%)        & ACC(\%)        & AUC(\%)        & AUC(\%)        \\ \hline
XceptionNet\cite{FF++}             & ICCV 2019              & 64.77          & 68.52          & 51.21          & 70.50          & 61.46          & 68.52          & 63.61          & 69.63          & 69.29          \\
F3-Net\cite{F3Net}                  & ECCV 2020              & \underline{72.99} & \underline{74.03}          & 65.09          & 84.13          & 63.55          & 72.13          & \underline{65.20}          & 67.76          & \underline{74.51}          \\
LTW\cite{LTW}                     & AAAI 2021              & 63.40          & 64.10          & 60.90          & 84.90          & 59.10          & 63.40          & 63.10          & 69.00          & 70.35          \\
MAT\cite{MAT}                     & CVPR 2021              & 70.79          & 73.87          & 65.59          & 83.78          & \underline{65.03} & \underline{73.33} & 61.96          & 65.88          & 74.22          \\
RECCE\cite{RECCE}                   & CVPR 2022              & 67.09          & 65.96          & 50.40          & 87.13          & 61.74          & 69.08          & 62.53          & 66.99          & 72.29          \\
CORE\cite{CORE}                    & CVPR 2022              & 68.78          & 68.67          & \textbf{71.41} & \underline{90.10}          & 62.75          & 68.04          & 61.45          & 67.36          & 73.54          \\
CADDM\cite{CADDM}                    & CVPR 2023              & 68.75          & 66.83          & 67.48         & 87.05          & 62.23          & 70.71          & 64.90          & 68.72          & 73.33          \\
IFFD\cite{IFFD}                    & TIP 2023               & 61.53          & 61.88          & 57.18          & 83.60          & 63.39          & 69.80          & 60.05          & 62.92          & 69.55          \\
UMFC\cite{UMFC}                    & AAAI 2024              & 69.63          & 70.68          & 56.72          & 83.44          & 64.62          & 69.81          & 63.71          & 68.64          & 73.14          \\
SFIConv\cite{SFIConv}                 & TIFS 2024              & 68.76          & 70.36          & 56.67          & 78.82          & 63.01          & 69.92          & 64.83          & 67.96          & 71.77          \\
DFGaze\cite{DFGaze}                  & TIFS 2024              & 58.99          & 72.52          & 53.09          & 78.09          & 61.34          & 67.42          & 57.09          & \underline{73.54} & 72.89          \\ \hline
FGL                    & Ours                   & \textbf{74.52}          & \textbf{77.14} & \underline{69.31}          & \textbf{93.75} & \textbf{65.94}          & \textbf{74.06}          & \textbf{68.69} & \textbf{74.14}          & \textbf{79.77} \\ 
\bottomrule
\end{tabular}
\end{table*}

\renewcommand{\arraystretch}{1.3}
\begin{table*}[t]
\small
\setlength{\tabcolsep}{1mm}
\centering
\caption{THE EVALUATION RESULTS ON UNKNOWN FORGERY TECHNIQUES}
\label{tab:table3}
\begin{tabular}{ccccccccccc}
\toprule
\multirow{2}{*}{Method} & \multirow{2}{*}{Venue} & \multicolumn{2}{c}{TEST-DF}     & \multicolumn{2}{c}{TEST-F2F}    & \multicolumn{2}{c}{TEST-FS}     & \multicolumn{2}{c}{TEST-NT}     & \multirow{2}{*}{\begin{tabular}[c]{@{}c@{}}Average\\ AUC(\%)\end{tabular}} \\ \cline{3-10}
                        &                        & ACC(\%)            & AUC(\%)            & ACC(\%)            & AUC(\%)            & ACC(\%)            & AUC(\%)            & ACC(\%)            & AUC(\%)            &                 \\ \hline
F3-Net\cite{F3Net}                  & ECCV 2020              & 82.31    & 89.50          & 63.85          & 70.00          & 49.68          & 54.04          & 57.85          & 63.29          & 69.21          \\
MAT\cite{MAT}                     & CVPR 2021              & 80.13          & 88.32          & 62.51          & 69.47          & 51.88          & 53.98          & 56.61          & 66.19          & 69.49          \\
LTW\cite{LTW}                  & AAAI 2021              & 77.00               & 87.42                & 63.86                & 74.22               & 51.22               & 55.27               & 53.68               & 64.32               & 70.31          \\ 
LTW (5-shot)                  & AAAI 2021              & 78.23               & 87.37                & \underline{66.98}                & 72.69               & 52.43               & \underline{56.34}               & 55.94               & 67.41               & 70.95          \\
CORE\cite{CORE}                    & CVPR2022               & 75.13          & 85.08          & 66.53          & \underline{75.86}    & 49.92          & 52.41          & 58.65          & 66.50          & 69.96          \\

CADDM\cite{CADDM}                   & CVPR 2023              & 70.39          & 84.92          & 59.56          & 68.25          & 50.85          & 52.90          & 55.36          & 64.26          & 67.58          \\
IFFD\cite{IFFD}                    & TIP 2023               & 82.26          & 90.38          & 66.85    & 73.75          & 50.45          & 53.98          & \underline{59.87}    & 66.91          & 71.26          \\
SFIConv\cite{SFIConv}                 & TIFS 2024               & 75.08          & 90.04          & 63.53          & 71.56          & 50.72          & 55.81          & 58.70          & 63.72          & 70.38          \\
DFGaze\cite{DFGaze}                  & TIFS 2024              & \underline{82.52}               & 91.13                & 63.73                & 70.94               & 51.57               & 55.66               & 57.44               & 65.82               & 70.89          \\
   \hline
FGL                    & Ours                   & \textbf{83.13} & \underline{91.83} & 63.50          & 75.45          & \underline{52.57}    & 55.35    & 54.57          & \underline{67.47}    & \underline{72.53}    \\ \hline
FGL (5-shot)            & Ours                   & 81.25          & \textbf{91.97}    & \textbf{71.27} & \textbf{78.29} & \textbf{55.25} & \textbf{59.15} & \textbf{64.16} & \textbf{70.53} & \textbf{74.64} \\ 
\bottomrule
\end{tabular}
\end{table*}

\subsection{Scenario 1: Evaluation on Unknown Forgery Datasets}
With the increasing number of deepfake video datasets and the significant differences between them, achieving good performance within a specific dataset is not enough. The primary challenge in deepfake face detection is achieving effective generalization on unseen datasets. Our proposed method is designed to achieve effective detection in cross-dataset scenarios. To verify the effectiveness of the proposed method, we train the model on FF++ (c23) and evaluate its performance on different datasets, including CDF, DFR, WDF, and DFDC. 

Table~\ref{tab:table2} summarizes the results of cross-dataset evaluation. In this experiment, the same settings are applied as those used in the comparison method, and no additional support set is utilized. Our proposed method demonstrates a clear advantage over others, with significantly better average results across the four datasets. Methods like RECCE\cite{RECCE} and LTW\cite{LTW} perform well on the FF++ dataset. However, they focus too much on specific forgery techniques, leading to poor performance in cross-dataset evaluations. Our proposed method achieves a balance between the results of the evaluation in the intra-dataset and in the cross-dataset. Notably, the F3-Net\cite{F3Net} and MAT\cite{MAT} methods perform well on the CDF and WDF datasets but show poor results on the other two datasets. For our method, there is no bias toward any specific dataset because it guides the model parameters to values that perform well across all datasets. However, our method does not achieve the best results on the DFDC dataset. We speculate that the forgery features in the DFDC dataset differ significantly from those in FF++, which prevents our method from finding parameters that generalize well to this dataset.

\subsection{Scenario 2: Evaluation on Unknown Forgery Techniques}
In the real world, new forgery techniques are constantly emerging, posing significant challenges to deepfake detection tasks. A good detection method must maintain strong performance when faced with unknown forgery techniques. In this section, we simulate real-world scenarios by evaluating the model on unknown forgery techniques. Specifically, we simulate real-world scenarios by excluding one of the four forgery methods from the FF++ dataset in each experiment to represent a newly emerging forgery technique. 

Table~\ref{tab:table3} presents the experimental results for evaluations on different unknown forgery techniques. We can draw a conclusion that most methods can't be well detected when faced with unknown forgery methods. The model performs well when detecting forgery methods like DF, which are older and exhibit obvious forgery traces. However, its performance significantly decreases when dealing with newer methods like FS and NT, where the forgery traces are subtle. More importantly, there are great differences between different deepfake techniques. Our method uses a small amount of data to learn this difference and improve the ability of the model to detect unknown forgery techniques. In the experiments, we present two versions of our method: one without using the support set and the other using the support set (5-shot). In FGL, we achieved good performance using only our DPNet. In FGL (5-shot), we apply our proposed FGL strategy to learn the unknown forgery techniques in the 5-shot support dataset, achieving optimal performance. 

Due to the limited availability of open-source meta-learning methods, we selected LTW\cite{LTW} for comparison in our experiments. LTW\cite{LTW} uses virtual updates during the meta-test phase, but the model’s learning capacity has not been significantly enhanced, resulting in average performance in detecting unknown forgery techniques. In contrast, FGL effectively guides the model to adapt to unknown forgery techniques. When tested on the F2F, FS, and NT datasets, it achieved significant improvements, with the accuracy on the NT dataset increasing by nearly 10\%. Therefore, we conclude that our method performs well in real-world scenarios and effectively addresses emerging forgery techniques.

\subsection{Ablation Study}
In this section, we conducted comprehensive ablation experiments to thoroughly evaluate our proposed method. Specifically, our ablation study includes the following components: 1) evaluation of the overall effectiveness of the proposed method; 2) demonstration of the effectiveness of the Forgery Guided Learning (FGL) strategy. 
\renewcommand{\arraystretch}{1.3}
\begin{table*}[htbp]
\centering
\caption{EVALUATION OF THE EFFECTIVENESS OF DIFFERENT COMPONENTS}
\label{tab:table4}
\begin{tabular}{cclclclclcl}
\toprule
\multirow{2}{*}{Method} & \multicolumn{2}{c}{FF++→CDF} & \multicolumn{2}{c}{FF++→DFR} & \multicolumn{2}{c}{FF+→WDF} & \multicolumn{2}{c}{FF+→DFDC} & \multicolumn{2}{c}{Average} \\ \cline{2-11} 
                        & \multicolumn{2}{c}{AUC(\%)}      & \multicolumn{2}{c}{AUC(\%)}      & \multicolumn{2}{c}{AUC(\%)}     & \multicolumn{2}{c}{AUC(\%)}      & \multicolumn{2}{c}{AUC(\%)}     \\ \hline
Baseline                & \multicolumn{2}{c}{67.64}    & \multicolumn{2}{c}{90.65}    & \multicolumn{2}{c}{68.26}   & \multicolumn{2}{c}{67.94}    & \multicolumn{2}{c}{73.62}   \\
w/ FPM                 & \multicolumn{2}{c}{68.6}     & \multicolumn{2}{c}{92.87}    & \multicolumn{2}{c}{69.18}   & \multicolumn{2}{c}{73.76}    & \multicolumn{2}{c}{76.10}   \\
w/ AFRP                & \multicolumn{2}{c}{73.64}    & \multicolumn{2}{c}{92.21}    & \multicolumn{2}{c}{71.53}   & \multicolumn{2}{c}{71.53}    & \multicolumn{2}{c}{77.22}   \\ \hline
DPNet (FPM+AFRP)        & \multicolumn{2}{c}{77.14}    & \multicolumn{2}{c}{93.75}    & \multicolumn{2}{c}{74.06}   & \multicolumn{2}{c}{74.14}    & \multicolumn{2}{c}{79.77}   \\ 
\bottomrule
\end{tabular}
\end{table*}

\renewcommand{\arraystretch}{1.3}
\begin{table}[ht]
\centering
\caption{EVALUATION OF THE EFFECTIVENESS OF FGL STRATEGY}
\label{tab:table5}
% \resizebox{\linewidth}{!}{
\begin{tabular}{ccccc}
\toprule
\multirow{2}{*}{Method} & \multicolumn{2}{c}{TEST-F2F} & \multicolumn{2}{c}{TEST-NT} \\ \cline{2-5} 
                        & ACC(\%)       & AUC(\%)      & ACC(\%)      & AUC(\%)          \\ \hline
w/o FGL                & 63.50         & 75.45        & 54.57        & 67.47        \\ \hline
w/ FGL (1-shot)         & 70.05         & 78.56        & 61.00        & 72.98        \\ \hline
w/ FGL (5-shot)         & 71.27         & 78.29        & 64.16        & 70.53        \\ 
\bottomrule
\end{tabular}
\end{table}

\subsubsection{Effectiveness of Different Components} To investigate the contribution of each component in our proposed method, we conducted a series of ablation experiments. Our method primarily relies on the Forgery Guided Learning (FGL) strategy and the Dual Perception Network (DPNet) for forgery detection. However, since FGL strategy functions as an additional learning process and inherently strengthens the model's generalization ability, this section focuses on evaluating the impact of DPNet. Specifically, we ablate the Frequency-domain Perception Mechanism (FPM) and the Adaptive Forgery Relationship Perception (AFRP) within the DPNet to assess their individual contributions. The experiments were trained on FF++ and tested on the other four datasets, with the AUC results shown in Table~\ref{tab:table4}. The FPM integrates frequency-domain information into our detection framework, enabling dynamic extraction of relevant features across different frequency components. With the incorporation of the AFRP module, the model's performance is significantly enhanced, demonstrating the effectiveness of our adaptive graph convolution in emphasizing essential feature information. Therefore, the proposed framework achieves robust performance in the designed method.

\subsubsection{Effectiveness of FGL Strategy} The core of FGL strategy is to guide the model to adapt to unknown forgery methods using few-shot forgery data. We evaluate our strategy in three configurations: without FGL, FGL (1-shot) and FGL (5-shot). The experiments were conducted on different forgery methods within the FF++ dataset, with the detailed results shown in Table~\ref{tab:table5}. We take F2F and NT, two unknown forgery techniques that are more difficult to detect, as our test set. Experiments show that our FGL strategy can detect unknown forgery techniques more effectively with only 1-shot data. Specifically, the FGL (1-shot) configuration achieved an average detection accuracy of 70.05\% on F2F and 61.00\% on NT, outperforming the baseline by 6.55\% and 6.43\% respectively. The performance gap between 1-shot and 5-shot configurations is small, demonstrating the efficiency of our few-shot learning approach. These results validate that our FGL strategy significantly enhances the model's adaptability to novel forgery techniques.

\begin{figure*}[htbp] % 使用 figure* 横跨两栏
    \centering
    \includegraphics[width=\textwidth]{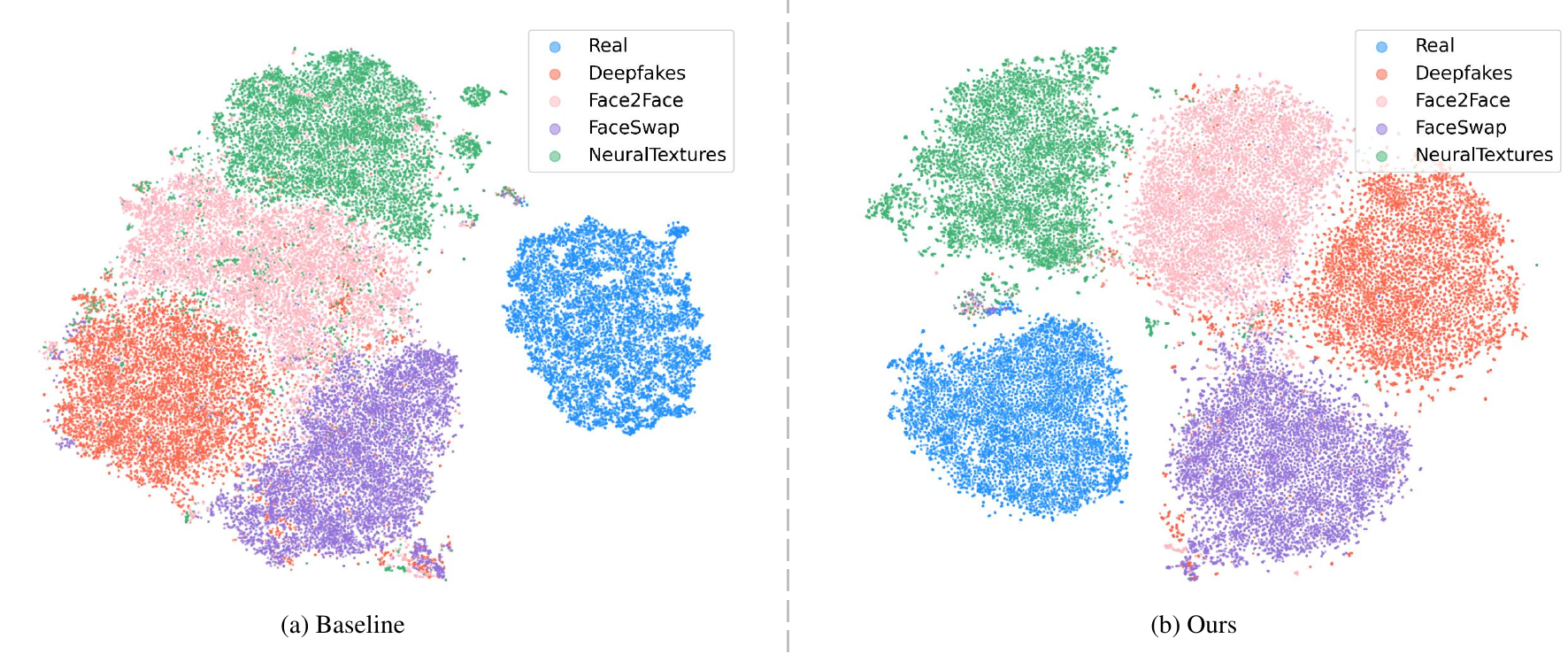} % 设置图片宽度为 \textwidth（页面宽度）
    \caption{T-SNE feature distribution visualizations of Baseline and our FGL. Our method can not only distinguish real images from forged images, but also learn the differences between different forgery techniques.}
    \label{fig:4}
\end{figure*}

\begin{figure}[htbp]
    \centering
    \includegraphics[width=0.6\columnwidth]{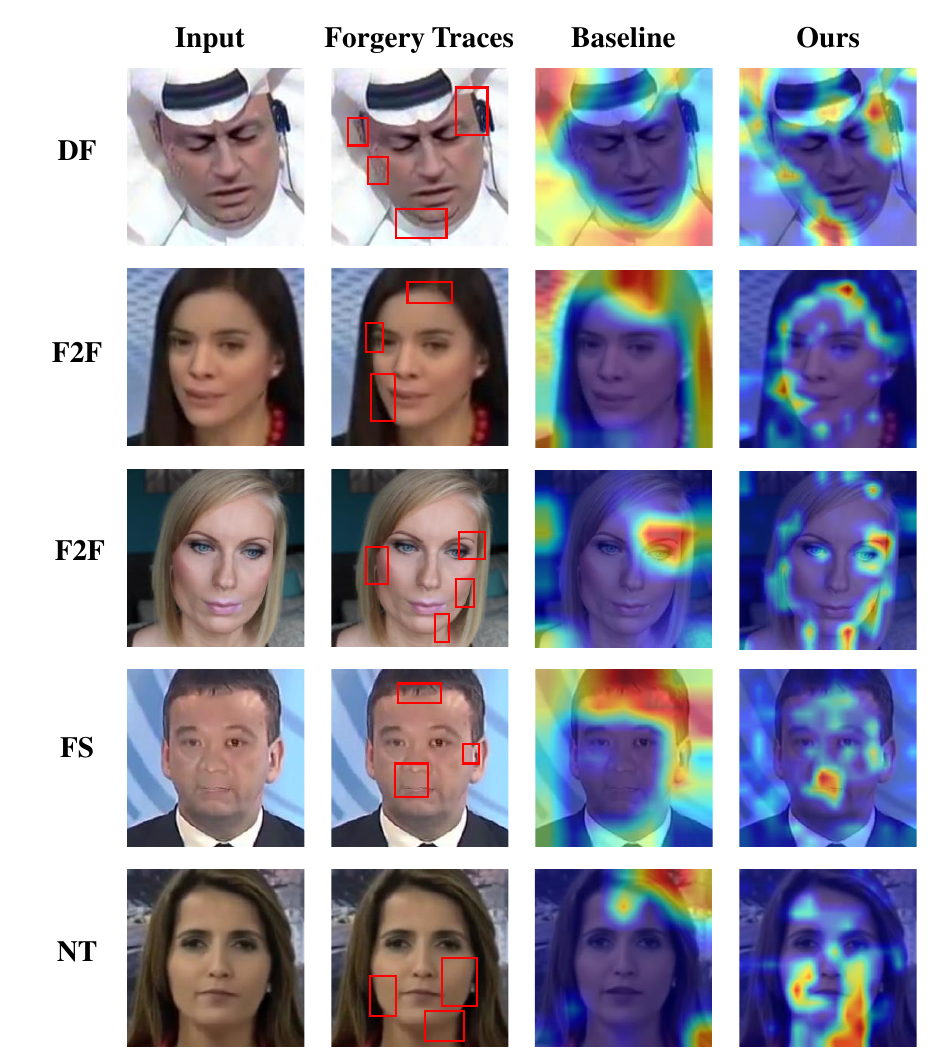} % 设置图片宽度为 \textwidth（页面宽度）
    \caption{Saliency map visualization of Baseline and our FGL. The red box indicates some obvious or subtle forgery traces.}
    \label{fig:5}
\end{figure}

\subsection{Visualization}
In this section, we present visualizations of the input feature distribution and saliency map to demonstrate the effectiveness of our FGL detection method.

\subsubsection{T-SNE Feature Embedding Visualization} We use t-SNE \cite{tsne} to visualize the input feature distributions for both the Baseline and our proposed method. The visualization results on the FF++ \cite{FF++} dataset are shown in Fig. \ref{fig:4}. Since different forgery methods exhibit distinct manipulation artifacts, our approach needs to not only distinguish real images but also differentiate between various forgery techniques. These differences guide the model to learn and adapt to unknown forgery techniques. The visualization results clearly demonstrate that our method achieves better inter-class separation compared to the baseline. Particularly for challenging forgery methods like DeepFakes\cite{DF} and Face2Face\cite{F2F}, our method shows distinct feature distribution patterns that are well-separated from both real images and other manipulation types. This improved feature discrimination ability directly contributes to the model's generalization ability in different forgery technologies. These visual observations align with our quantitative results, providing additional evidence for the superiority of our approach.

\subsubsection{Saliency Map Visualization} We use Grad-CAM \cite{gradcam} to highlight the regions that our proposed method focuses on in forged faces. Fig. \ref{fig:5} presents the visualization results on FF++ dataset, demonstrating that our method pays more attention to subtle forgery clues. It is important to note that different forgery methods leave distinct traces: DF and F2F typically exhibit clues focused on facial contours and edges, while FS and NT highlight artifacts related to facial expression muscles and movements. Our visualization results reveal that FGL can accurately capture these subtle forgery features, whereas the Baseline method often only identifies obvious manipulation traces. Especially when dealing with NT, our method can identify subtle texture-level differences that are usually challenging for traditional methods to detect. This refined attention distribution enables our method to demonstrate stronger adaptability and robustness when handling different forgery techniques.

\section{Conclusion}
\label{sec5}
In this study, we focus on enhancing the generalization capability of detection models against unknown forgery techniques. To address the challenges posed by the rapid evolution of deepfake technologies, we propose the Forgery Guided Learning (FGL) strategy. The FGL enables the detection model to dynamically adapt to unknown forgery techniques while maintaining stable performance, even as forgery techniques continue to evolve.
Furthermore, we integrate a Frequency-domain Perception Mechanism (FPM) and an Adaptive Forgery Relationship Perception (AFRP) module to construct a Dual Perception Network (DPNet). This network improves the ability of the model to capture forgery traces and optimizes feature interactions. By combining FGL strategy with DPNet, our approach improves the precision of forgery identification.
Extensive experimental results demonstrate that FGL significantly outperforms existing methods in handling unknown forgery techniques. This highlights its effectiveness in real-world deepfake detection scenarios. In future work, we aim to further extend this framework to accommodate more diverse and complex forgery scenarios.

%Bibliography
\bibliographystyle{unsrt}  
\bibliography{references}

\end{document}